\documentclass[letterpaper, 10 pt, conference]{ieeeconf}  

\IEEEoverridecommandlockouts                              

\usepackage{graphics} 
\usepackage{graphicx}
\usepackage{algpseudocode,algorithm,algorithmicx}
\usepackage{rotating}
\usepackage{color}
\usepackage{tikz}
\usetikzlibrary{plotmarks}
\usepackage{hyperref}
\usetikzlibrary{arrows.meta}

\algrenewcommand\algorithmicrequire{\textbf{Precondition:}}
\algrenewcommand\algorithmicensure{\textbf{Postcondition:}}
\usepackage{tabularx,booktabs}
\newcolumntype{C}{>{\centering\arraybackslash}X} 
\usepackage{float}
\usepackage{subfig}
\usepackage{multirow}
\usepackage{pgfplots}
\usepackage{amsmath}

\title{\LARGE \bf DynaSLAM: Tracking, Mapping and Inpainting in Dynamic Scenes} 

\author{Berta Bescos, Jos\'e M. F\'acil, Javier Civera and Jos\'e Neira
\thanks{This work has been supported by NVIDIA Corporation through the donation of a Titan X GPU, by the Spanish Ministry of Economy and Competitiveness (projects DPI2015-68905-P and DPI2015-67275-P, FPI grant BES-2016-077836), and by the Arag\'on regional government (Grupo DGA T04-FSE).}
\thanks{Berta Bescos, Jos\'e M. F\'acil, Javier Civera and Jos\'e Neira are with the Instituto de Investigaci\'on en Ingenier\'ia de Arag\'on (I3A), Universidad de Zaragoza, Zaragoza 50018, Spain {\tt\small \{bbescos,jmfacil,jcivera,jneira\}@unizar.es}}%
}
\begin{document}

\maketitle
\thispagestyle{empty} 
\pagestyle{empty} 

\begin{abstract} 
The assumption of scene rigidity  is typical in SLAM algorithms. Such a strong assumption limits the use of most visual SLAM systems in populated real-world environments, which are the target of several relevant applications like service robotics or autonomous vehicles.

In this paper we present DynaSLAM, a visual SLAM system that, building on ORB-SLAM2 \cite{mur2017orb}, adds the capabilities of dynamic object detection and background inpainting. DynaSLAM is robust in dynamic scenarios for monocular, stereo and \mbox{RGB-D} configurations. We are capable of detecting the moving objects either by multi-view geometry, deep learning or both. Having a static map of the scene allows inpainting the frame background that has been occluded by such dynamic objects.

We evaluate our system in public monocular, stereo and \mbox{RGB-D} datasets. We study the impact of several accuracy/speed trade-offs to assess the limits of the proposed methodology. DynaSLAM outperforms the accuracy of standard visual SLAM baselines in highly dynamic scenarios. And it also estimates a map of the static parts of the scene, which is a must for long-term applications in real-world environments.
\end{abstract}


\section{INTRODUCTION}
Simultaneous Localization and Mapping (SLAM) is a prerequisite for many robotic applications, for example collision-less navigation. SLAM techniques estimate jointly a map of an unknown environment and the robot pose within such map, only from the data streams of its on-board sensors. The map allows the robot to continually localize within the same environment without accumulating drift. This is in contrast to odometry approaches that integrate the incremental motion estimated within a local window and are unable to correct the drift when revisiting places. 

Visual SLAM, where the main sensor is a camera, has received a high degree of attention and research efforts over the last years. The minimalistic solution of a monocular camera has practical advantages with respect to size, power and cost, but also several challenges such as the unobservability of the scale or  state initialization. By using more complex setups, like stereo or \mbox{RGB-D} cameras, these issues are solved and the robustness of visual SLAM systems can be greatly improved.

\begin{figure} [t]
\centering
\subfloat[\label{fig:in} Input \mbox{RGB-D} frames with dynamic content.] {\includegraphics[width=1\linewidth]{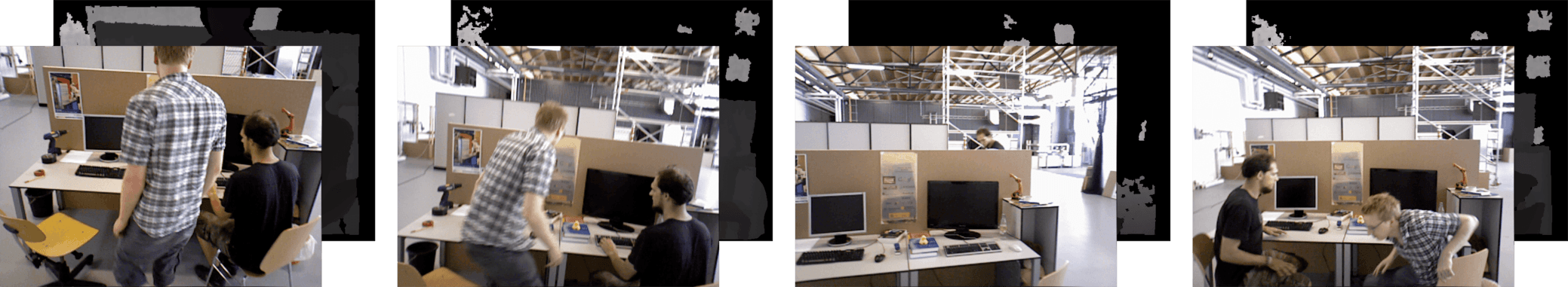}} \\
\subfloat[\label{fig:outFrames} Output \mbox{RGB-D} frames. Dynamic content has been removed. Occluded background has been reconstructed with information from previous views.] {\includegraphics[width=1\linewidth]{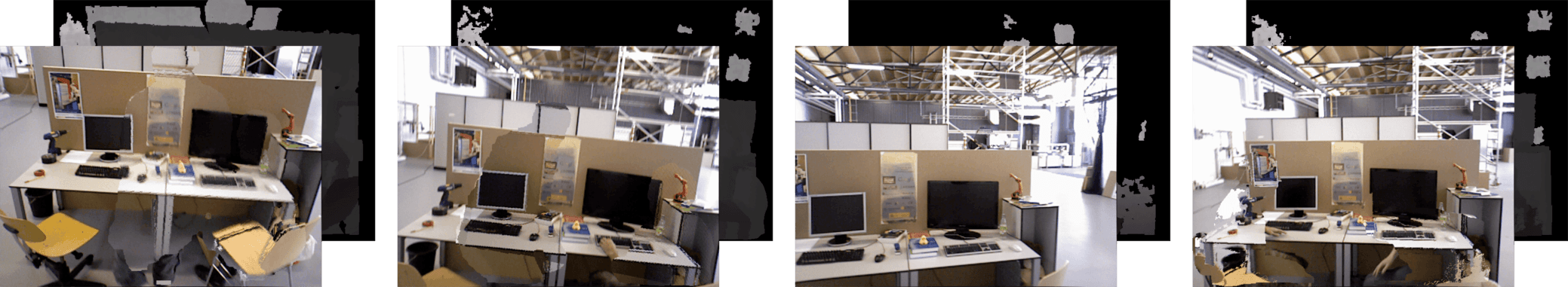}} \\
\subfloat[\label{fig:outPC} Map of the static part of the scene, after removal of the dynamic objects.] {\includegraphics[width=1\linewidth]{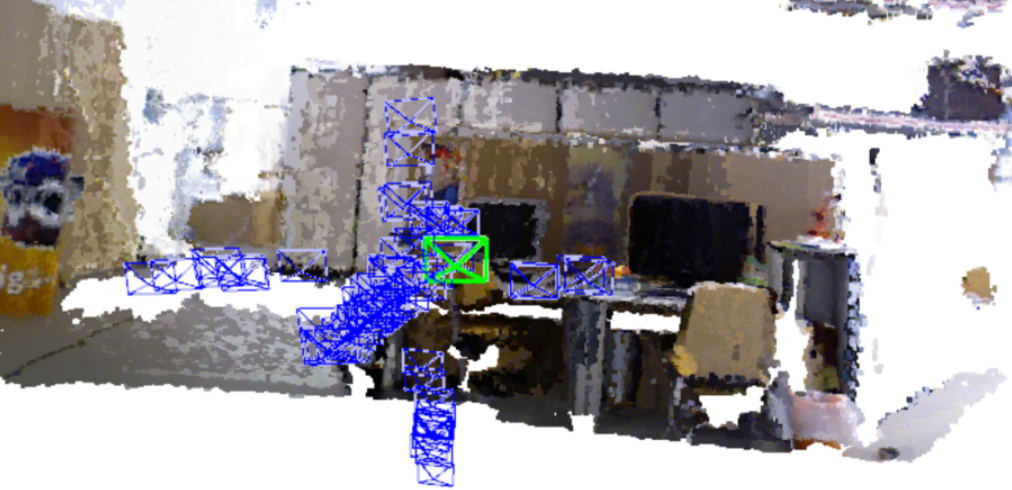}} \\
\caption{\label{fig:overview} Overview of DynaSLAM results for the \mbox{RGB-D} case.}
\end{figure}

The research community has addressed SLAM from many different angles. However, the vast majority of the approaches and datasets assume a static environment. As a consequence, they can only manage small fractions of dynamic content by classifying them as outliers to such static model. Although the static assumption holds for some robotic applications, it limits the applicability of visual SLAM in many relevant cases, such as intelligent autonomous systems operating in populated real-world environments over long periods of time.


Visual SLAM can be classified into feature-based methods \cite{klein2007parallel,mur2015orb}, that rely on salient points matching and can only estimate a sparse reconstruction; and direct methods \cite{stuhmer2010real,newcombe2011dtam,graber2011online}, which are able to estimate in principle a completely dense reconstruction by the direct minimization of the photometric error and TV regularization. Some direct methods focus on the high-gradient areas estimating semi-dense maps \cite{engel2014lsd,engel2017direct}. 

None of the above methods, considered the state of the art, address the very common problem of dynamic objects in the scene, \textit{e.g.}, people walking, bicycles or cars. Detecting and dealing with dynamic objects in visual SLAM reveals several challenges for both mapping and tracking, including:

\begin{enumerate}
\item
How to detect such dynamic objects in the images to:
\begin{enumerate}
\item
Prevent the tracking algorithm from using matches that belong to dynamic objects.
\item
Prevent the mapping algorithm from including moving objects as part of the 3D map.
\end{enumerate}
\item
How to complete the part of the 3D map that is temporally occluded by a moving object.
\end{enumerate}

Many applications would greatly benefit from progress along these lines.  Among others, augmented reality, autonomous vehicles, and medical imaging. All of them could for instance safely reuse maps from previous runs. 
Detecting and dealing with dynamic objects is a requisite to estimate stable maps, useful for long-term applications. If the dynamic content is not detected, it becomes part of the 3D map, complicating its usability for tracking or relocation purposes.

In this work we propose an on-line algorithm to deal with dynamic objects in \mbox{RGB-D}, stereo and monocular SLAM. 
This is done by adding a front-end stage to the state-of-the-art ORB-SLAM2 system \cite{mur2017orb}, with the purpose of having a more accurate tracking and a reusable map of the scene. 
In the monocular and stereo cases our proposal is to use a CNN to pixel-wise segment the \textit{a priori} dynamic objects in the frames (\textit{e.g.}, people and cars), so that the SLAM algorithm does not extract features on them. 
In the RGB-D case we propose to combine multi-view geometry models and deep-learning-based algorithms for detecting dynamic objects and, after having removed them from the images, inpaint the occluded background with the correct information of the scene (Fig. \ref{fig:overview}).

The rest of the paper is structured as follows: section \ref{sec:related} discusses related work, section \ref{sec:system} gives the details of our proposal, section \ref{sec:experiments} details the experimental results, and section \ref{sec:conclusions} presents the conclusions and lines for future work.

\section{RELATED WORK}
\label{sec:related}
Dynamic objects are, in most SLAM systems, classified as spurious data and therefore neither included in the map nor used for camera tracking. The most typical outlier rejection algorithms are RANSAC (\textit{e.g.}, in ORB-SLAM \cite{mur2015orb}, \cite{mur2017orb}) and robust cost functions (\textit{e.g.}, in PTAM \cite{klein2007parallel}).

There are several SLAM systems that address more specifically the dynamic scene content. Within  feature-based SLAM methods, some of the most relevant are:

\begin{itemize}

\item Tan \textit{et al.} \cite{tan2013robust} detect changes that take place in the scene by projecting the map features into the current frame for appearance and structure validation. 

\item Wangsiripitak and Murray \cite{wangsiripitak2009avoiding} track known 3D dynamic objects in the scene. Similarly, Riazuelo \textit{et~al.} \cite{riazuelo2017semantic} deal with human activity by detecting and tracking people.
%

\item More recently, the work of Li and Lee \cite{li2017rgb} uses depth edges points, which have an associated weight indicating its probability of belonging to a dynamic object.

\end{itemize}

Direct methods are, in general, more sensitive to dynamic objects in the scene. The most relevant works specifically designed for dynamic scenes are:

\begin{figure*} [ht!]
\centering
\includegraphics[width=1\linewidth]{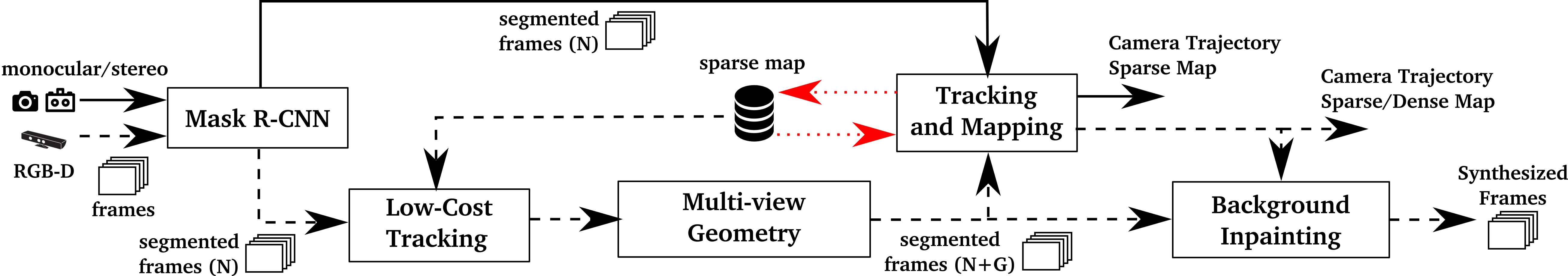}
\caption{\label{fig:diagram}Block diagram of our proposal. In the stereo and monocular pipeline (black continuous line) the images pass through a Convolutional Neural Network (\mbox{Mask R-CNN}) for computing the pixel-wise semantic segmentation of the \textit{a priori} dynamic objects before being used for the mapping and tracking. In the \mbox{RGB-D} case (black dashed line) a second approach based on multi-view geometry is added for a more accurate motion segmentation, for which we need a low-cost tracking algorithm. Once the position of the camera is known (Tracking and Mapping output), we can inpaint the background occluded by dynamic objects. The red dotted line represents the data flow of the stored sparse map.}
\end{figure*}

\begin{itemize}
\item
Alcantarilla \textit{et al.} \cite{alcantarilla2012combining} detect moving objects by means of a scene flow representation with stereo cameras.
\item
Wang and Huang \cite{wang2014motion} segment the dynamic objects in the scene using RGB optical flow.
\item
Kim \textit{et~al.} \cite{kim2016effective} propose to obtain the static parts of the scene by computing the difference between consecutive depth images projected over the same plane.
\item
Sun \textit{et~al.} \cite{sun2017improving} calculate the difference in intensity between consecutive RGB images. Pixel classification is done with the segmentation of the quantized depth image.
\end{itemize}

All the methods --both feature-based and direct ones-- that map the static scene parts only from the information contained in the sequence \cite{mur2017orb, mur2015orb, tan2013robust, li2017rgb, alcantarilla2012combining, wang2014motion, kim2016effective, sun2017improving, concha2015dpptam}, fail to estimate lifelong models when an \textit{a priori} dynamic object remains static, \textit{e.g.}, parked cars or people sitting. 
On the other hand, Wangsiripitak and Murray \cite{wangsiripitak2009avoiding}, and Riazuelo \textit{et al.} \cite{riazuelo2017semantic} would detect those \textit{a priori} dynamic objects, but would fail to detect changes produced by static objects, \textit{e.g.}, a chair a person is pushing, or a ball that someone has thrown. 
That is, the former approach succeeds in detecting \textit{moving} objects, and the second one in detecting several \textit{movable} objects. Our proposal, DynaSLAM, combines multi-view geometry and deep learning in order to address both situations. 
Similarly, Anrus \textit{et~al.} \cite{ambrus2016unsupervised} segment dynamic objects by combining a dynamic classifier and multi-view geometry.

\section{SYSTEM DESCRIPTION}
\label{sec:system}


Fig. \ref{fig:diagram} shows an overview of our system. 
First of all, the RGB channels pass through a CNN that segments out pixel-wise all the \textit{a priori} dynamic content, \textit{e.g.}, people or vehicles. 

In the \mbox{RGB-D} case, we use multi-view geometry to improve the dynamic content segmentation in two ways. 
First, we refine the segmentation of the dynamic objects previously obtained by the CNN. 
Second, we label as dynamic new object instances that are static most of the time (\textit{i.e.}, detect \textit{moving} objects that were not set to \textit{movable} in the CNN stage).

For that purpose, it is necessary to know the camera pose, for which a low-cost tracking module has been implemented to localize the camera within the already created scene map. 
These segmented frames are the ones which are used to obtain the camera trajectory and the map of the scene.
Notice that if the moving objects in the scene are not within the CNN classes, the multi-view geometry stage would still detect the dynamic content, but the accuracy might decrease.

Once this full dynamic object detection and localization of the camera have been done, we aim to reconstruct the occluded background of the current frame with static information from previous views. These synthetic frames are relevant for applications like augmented and virtual reality, and place recognition in lifelong mapping.

In the monocular and stereo cases, the images are segmented by the CNN so that keypoints belonging to the \textit{a priori} dynamic objects are neither tracked nor mapped. 

All the different stages are described in depth in the next subsections (\ref{subsec:FCN} to \ref{subsec:BR}).

\subsection{Segmentation of Potentially Dynamic Content using a CNN} 
\label{subsec:FCN}
For detecting dynamic objects we propose to use a CNN that obtains a pixel-wise semantic segmentation of the images. 
In our experiments we use \mbox{Mask R-CNN} \cite{he2017mask}, which is the state of the art for object instance segmentation. \mbox{Mask R-CNN} can obtain both pixel-wise semantic segmentation and the instance labels. 
For this work we use the pixel-wise semantic segmentation information, but the instance labels could be useful in future work for the tracking of the different moving objects. 
We use the TensorFlow implementation by Matterport\footnote{\href{https://github.com/matterport/Mask_RCNN}{\tt https://github.com/matterport/Mask\_RCNN}}.

The input of \mbox{Mask R-CNN} is the RGB original image. The idea is to segment those classes that are potentially dynamic or movable (person, bicycle, car, motorcycle, airplane, bus, train, truck, boat, bird, cat, dog, horse, sheep, cow, elephant, bear, zebra and giraffe). We consider that, for most environments, the dynamic objects likely to  appear are included within this list. If other classes were needed, the network, trained on \mbox{MS COCO} \cite{lin2014microsoft}, could be fine-tuned with new training data.

The output of the network, assuming that the input is an RGB image of size $m \times n \times 3$, is a matrix of size $m \times n \times l$, where $l$ is the number of objects in the image. For each output channel $i \in l$ a binary mask is obtained. By combining all the channels into one, we can obtain the segmentation of all dynamic objects appearing in one image of the scene.

\subsection{Low-Cost Tracking} 
\label{subsec:Light}
After the potentially dynamic content has been segmented, the pose of the camera is tracked using the static part of the image. Because the segment contours usually become high-gradient areas,  salient point features tend to appear. We do not consider the features in such contour areas.


The tracking implemented at this stage of the algorithm is a simpler and therefore computationally lighter version of the one in ORB-SLAM2 \cite{mur2017orb}. It projects the map features in the image frame, searches for the correspondences in the static areas of the image, and minimizes the reprojection error to optimize the camera pose. 


\subsection{Segmentation of Dynamic Content using \mbox{Mask R-CNN} and Multi-view Geometry} \label{subsec:FDOD}
By using \mbox{Mask R-CNN}, most of the dynamic objects can be segmented and not used for tracking and mapping. However, there are objects that cannot be detected by this approach because they are not \textit{a priori} dynamic, but movable. Examples of the latest are a book carried by someone, a chair that someone is moving, or even furniture changes in long-term mapping. The approach utilized for dealing with these cases is detailed in this section. 

\begin{figure}[b!]
\centering
\subfloat[\label{fig:est} Keypoint $x'$ belongs to a static object ($z' = z_{proj}$).] {\includegraphics[width=.43
\linewidth]{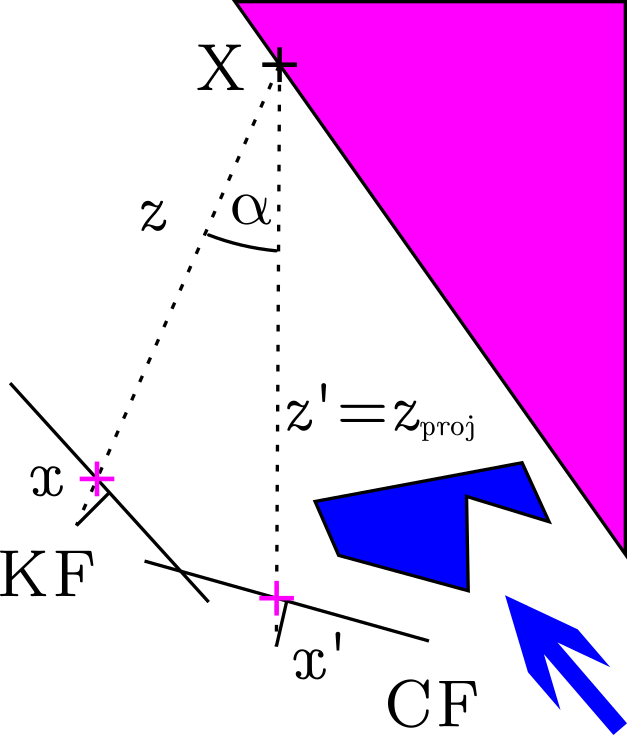}}
\hspace*{0.10\linewidth}
\subfloat[\label{fig:est} Keypoint $x'$ belongs to a dynamic object ($z' \ll z_{proj}$).] {\includegraphics[width=.43
\linewidth]{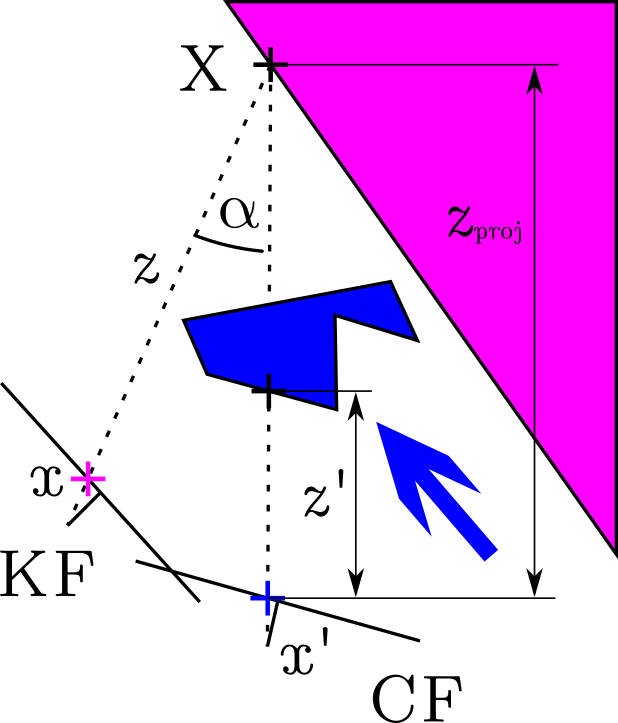}}
\caption{\label{fig:geo} Keypoint $x$ from the Key Frame (KF) is projected into the Current Frame (CF) using its depth and camera pose, resulting in point $x'$ with depth $z'$. The projected depth $z_{proj}$ is then computed. A pixel is labeled as dynamic if the difference $\Delta z = z_{proj} - z'$ is greater than a threshold $\tau_z$.} 
\end{figure}

\begin{figure*}[h!]
\centering 
\hspace{\fill}
\subfloat[\label{fig:geometric_seg} Using Multi-view Geometry.]{\includegraphics[width=.30
\linewidth]{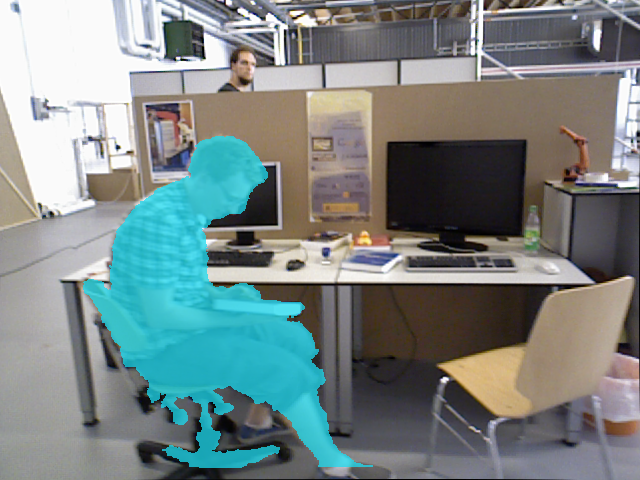}}
\hspace{\fill}
\subfloat[\label{fig:learning_seg} Using Deep Learning.]{\includegraphics[width=.30
\linewidth]{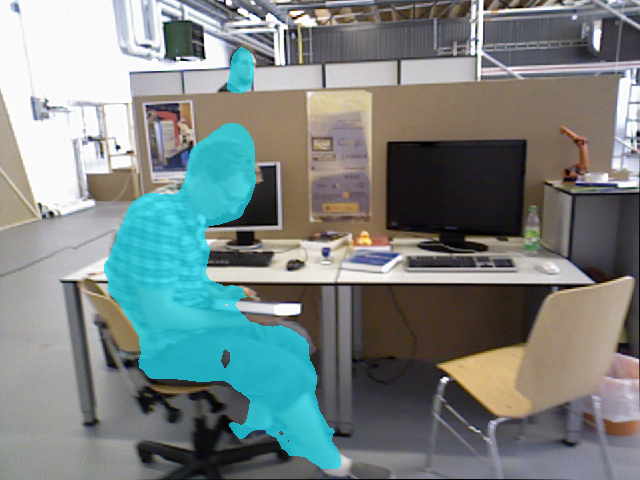}}
\hspace{\fill}
\subfloat[\label{fig:SegmentedFinal} Using Geometry and Deep Learning.]
{\includegraphics[width=.30\linewidth]{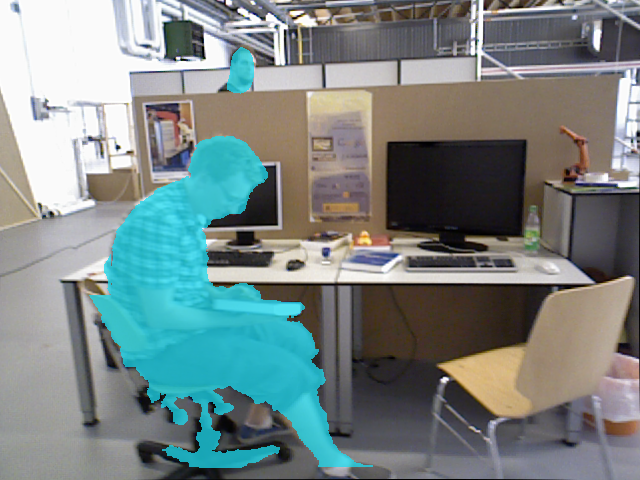}}
\hspace{\fill}
\caption{\label{fig:test1} Detection and segmentation of dynamic objects using multi-view geometry (left), deep learning (middle), and a combination of both geometric and learning methods (right). Notice that Fig.~\ref{fig:geometric_seg} cannot detect the person behind the desk, Fig.~\ref{fig:learning_seg} cannot segment the book carried by the person, and the combination of the two (Fig.~\ref{fig:SegmentedFinal}) is the best performing.}
\end{figure*}

For each input frame, we select the previous keyframes that have the highest overlaps. This is done by taking into account both the distance and the rotation between the new frame and each of the keyframes, similarly to Tan {\em et~al.} \cite{tan2013robust}. The number of overlapping keyframes has been set to $5$ in our experiments, as a compromise between computational cost and accuracy in the detection of dynamic objects.

We then compute the projection of each keypoint $x$ from the previous keyframes into the current frame, obtaining the keypoints $x'$, as well as their projected depth $z_{proj}$, computed from the camera motion.
Notice that the keypoints~$x$ come from the features extractor algorithm used in \mbox{ORB-SLAM2}.
For each keypoint, whose corresponding 3D point is $X$, we calculate the angle between the back-projections of $x$ and $x'$, \textit{i.e}., their parallax angle {$\alpha$}. 
If this angle is greater than 30$^{\circ}$, the point might be occluded, and will be ignored from then on.
We observed that, in the TUM dataset, for parallax angles greater than 30$^{\circ}$ static objects were considered as dynamic due to their viewpoint difference. 
We obtain the depth of the remaining keypoints in the current frame $z'$ (directly from the depth measurement), taking into account the reprojection error, and we compare them with $z_{proj}$. If the difference $\Delta z = z_{proj} - z'$ is over a threshold $\tau_z$, keypoint $x'$ is considered to belong to a dynamic object. This idea is shown in Fig. \ref{fig:geo}. 
To set the threshold $\tau_z$, we manually tagged the dynamic objects of 30 images within the TUM dataset, and evaluated both the precision and recall of our method for different thresholds $\tau_z$. By maximizing the expression $0.7 \times Precision + 0.3 \times Recall$, we concluded that $\tau_z = 0.4 m$ is a reasonable choice.


Some of the keypoints labeled as dynamic lay on the borders of moving objects, and might cause problems. To avoid this, we use the information given by the depth images. If a keypoint is set as dynamic, but a patch around itself in the depth map has high variance, we change the label to static.

So far, we know which keypoints belong to dynamic objects, and which ones do not. 
To classify all the pixels belonging to dynamic objects, we grow the region in the depth image around the dynamic pixels~\cite{gerlach2014evaluation}.
%
%
An example of a RGB frame and its corresponding dynamic mask can be seen in Fig. \ref{fig:geometric_seg}. 

The results of the CNN (Fig. \ref{fig:learning_seg}) can be combined with those of this geometric method for full dynamic object detection (Fig. \ref{fig:SegmentedFinal}). We can find strengths and limitations in both methods, hence the motivation for their combined use. For geometric approaches, the main problem is that initialization is not trivial because of its multi-view nature. Learning methods and their impressive performance using a single view, do not have such initialization problems. Their main limitation though is that objects that are supposed to be static can be moved, and the method is not able to identify them. This last case can be solved using multi-view consistency tests. 

These two ways of facing the moving objects detection problem are illustrated in Fig. \ref{fig:test1}. In Fig. \ref{fig:geometric_seg} we see that the person in the back, which is potentially a dynamic object, is not detected. There are two reasons for this. First, the difficulties that \mbox{RGB-D} cameras face when measuring the depth of distant objects. And second, the fact that reliable features lie on defined, and therefore nearby, parts of the image. Albeit, this person is detected by the deep learning method (Fig. \ref{fig:learning_seg}). Apart from this, on one hand we see in the Fig. \ref{fig:geometric_seg} that not only is detected the person in the front of the image, but also the book he is holding and the chair he is sitting on. On the other hand, in the Fig. \ref{fig:learning_seg} the two people are the only objects detected as dynamic, and also their segmentation is less accurate. If only the deep learning method is used, a \textit{floating book} would be left in the images and would incorrectly become part of the 3D map.

Because of the advantages and disadvantages of both methods, we consider that they are complementary and therefore their combined use is an effective way of achieving accurate tracking and mapping. In order to achieve this goal, if an object has been detected with both approaches, the segmentation mask should be that of the geometrical method. If an object has only been detected by the learning based method, the segmentation mask should contain this information too. The final segmented image of the example in the previous paragraph can be seen in the Fig. \ref{fig:SegmentedFinal}. The segmented dynamic parts are removed from the current frame and from the map.

\subsection{Tracking and Mapping} \label{subsec:TM}

The input to this stage of the system contains the RGB and depth images, as well as their segmentation mask. We extract ORB features in the image segments classified as static. As the segment contours are high-gradient areas, the keypoints falling in this intersection have to be removed. 

\begin{figure*}[h!]
\centering
\hspace{\fill}
\subfloat[\label{fig:inputSystemRGB} RGB original images.]{
\includegraphics[width=.23\linewidth]{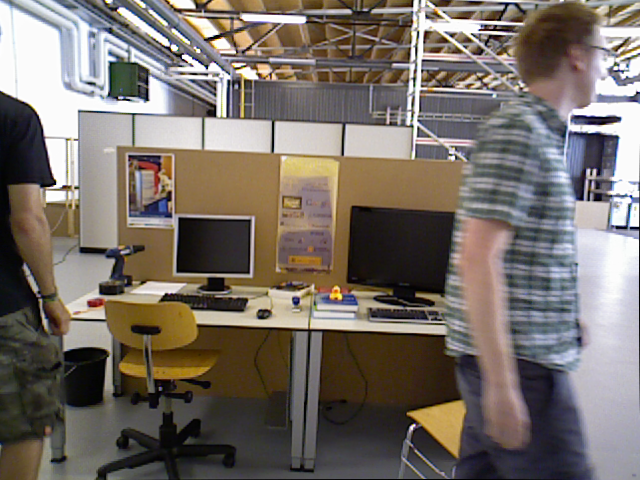}
\hspace{\fill}
\includegraphics[width=.23\linewidth]{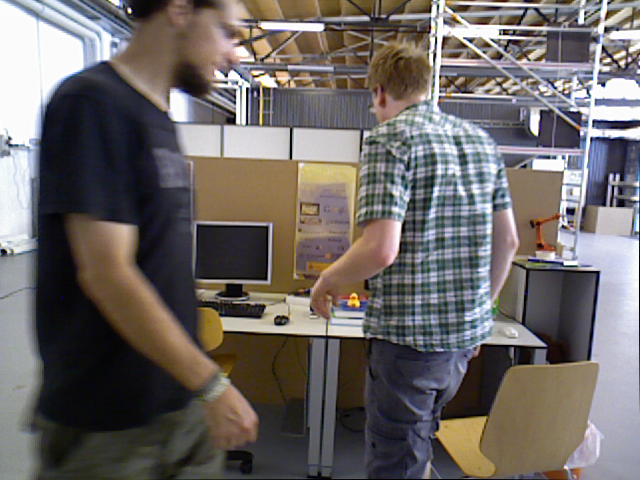}
\hspace{\fill}
\includegraphics[width=.23\linewidth]{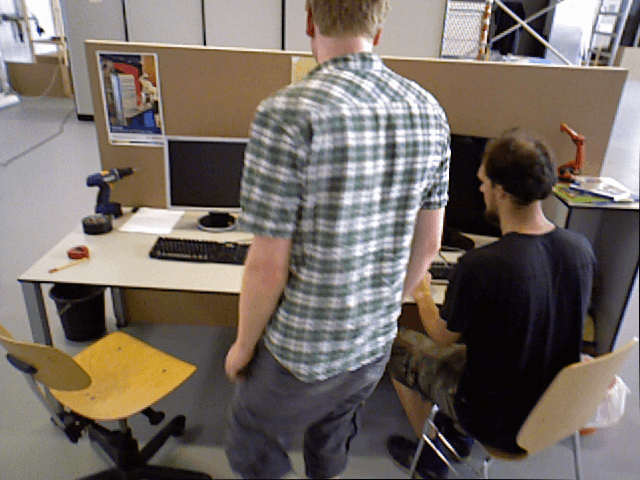} 
}
\hspace{\fill}
\subfloat[\label{fig:inputSystemDepth} Depth original image.]{
\includegraphics[width=.23\linewidth]{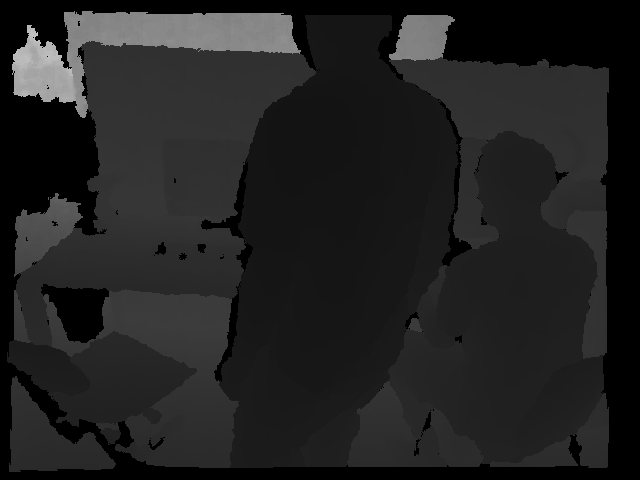}}
\hspace{\fill}
\hspace{\fill}
\\
\vspace*{-0.65\baselineskip}
\hspace{\fill}
\subfloat[\label{fig:outputSystemRGB} Inpainted RGB images.]{
\includegraphics[width=.23\linewidth]{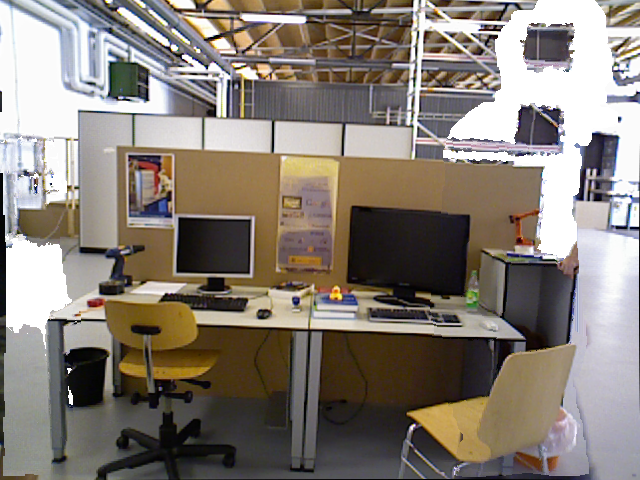}
\hspace{\fill}
\includegraphics[width=.23\linewidth]{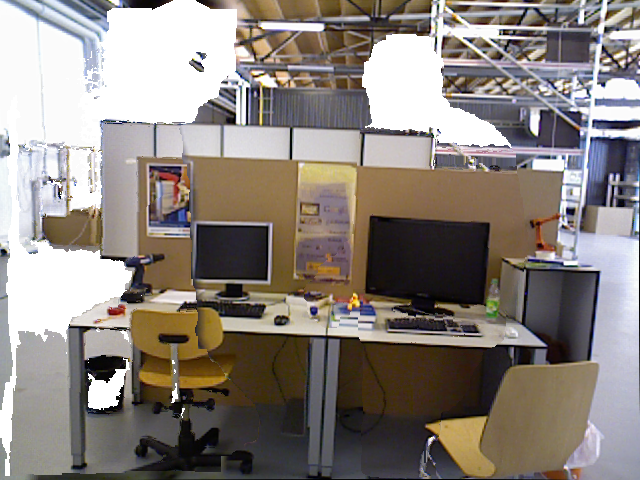}
\hspace{\fill}
\includegraphics[width=.23\linewidth]{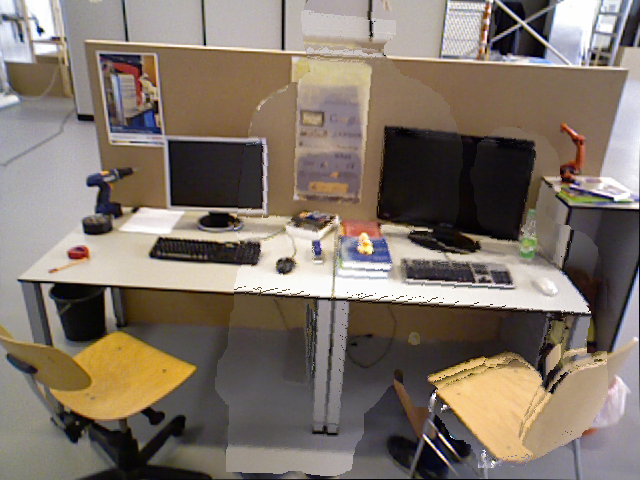} 
}
\hspace{\fill}
\subfloat[\label{fig:outputSystemDepth} Inpainted depth image.]{
\includegraphics[width=.23\linewidth]{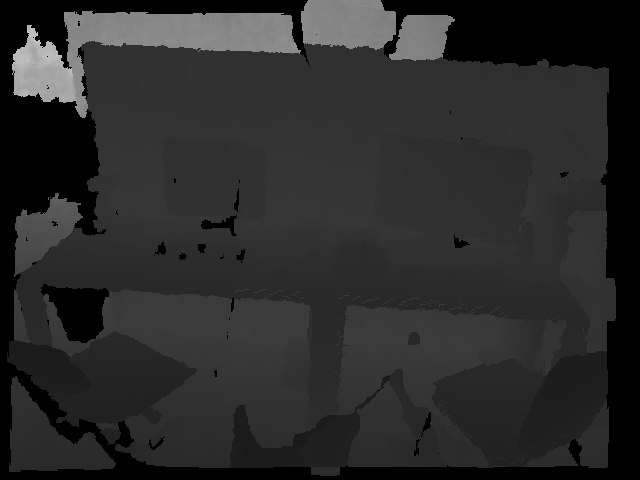}}
\hspace{\fill}
\vspace*{-0.2\baselineskip}
\caption{\label{fig:visual results} Qualitative results of our approach. In Fig. \ref{fig:inputSystemRGB} we show three RGB input frames, and in Fig. \ref{fig:outputSystemRGB} we show the output of our system, in which all dynamic objects have been detected and the background has been reconstructed. Figs. \ref{fig:inputSystemDepth} and \ref{fig:outputSystemDepth} show respectively the depth input and output, which has also been processed. Figure best viewed in electronic format.}
\end{figure*}

\subsection{Background Inpainting} \label{subsec:BR}

For every removed dynamic object, we aim at inpainting the occluded background with static information from previous views, so that we can synthesize a realistic image without moving content. 
We believe that such synthetic frames, containing the static structure of the environment, are useful for applications such as virtual and augmented reality, and for relocation and camera tracking after the map is created.

Since we know the position of the previous and current frames, we project into the dynamic segments of the current frame the RGB and depth channels from a set of all the previous keyframes (the last 20 in our experiments).
Some gaps have no correspondences and are left blank: some areas cannot be inpainted because their correspondent part of the scene has not appeared so far in the keyframes, or, if it has appeared, it has no valid depth information. 
These gaps cannot be reconstructed with geometrical methods and would need a more elaborate inpainting technique.
Fig. \ref{fig:visual results} shows the resulting synthetic images for three input frames from different sequences of the TUM benchmark. 
Notice how the dynamic content has been successfully segmented and removed. 
Also, most of the segmented parts have been properly inpainted with information from the static background. 

Another application of these synthesized frames would be the following: if the frames dynamic areas are inpainted with the static content, the system can work as a SLAM system under the staticity assumption using the inpainted images.

\section{EXPERIMENTAL RESULTS}
\label{sec:experiments}

We have evaluated our system in the public datasets TUM RGB-D and KITTI and compared to other state-of-the-art SLAM systems in dynamic environments, using when possible results published in the original papers. Furthermore we have compared our system against the original ORB-SLAM2 to quantify the improvement of our approach in dynamic scenes. In this case, the results for some sequences were not published and we have ourselves completed their evaluation. Mur and Tard\'os \cite{mur2017orb} propose to run each sequence five times and show median results, to account for the non-deterministic nature of the system. We have run each sequence ten times, as dynamic objects are prone to increase this non-deterministic effect.

\subsection{TUM Dataset}
The TUM \mbox{RGB-D} dataset \cite{sturm2012benchmark} is composed of 39 sequences recorded with a Microsoft Kinect sensor in different indoor scenes at full frame rate (30Hz). Both the RGB and the depth images are available, together with the ground-truth trajectory, the latest recorded by a high-accuracy motion-capture system. In the sequences named \textit{sitting} (\textit{s}) there are two people sitting in front of a desk while speaking and gesticulating, \textit{i.e.}, there is a low degree of motion. In the sequences named \textit{walking} (\textit{w}), two people walk both in the background and the foreground and sit down in front of the desk. This dataset is highly dynamic and therefore challenging for standard SLAM systems. For both types of sequences \textit{sitting} (\textit{s}) and \textit{walking} (\textit{w}) there are four types of camera motions: (1) halfsphere (half): the camera moves following the trajectory of a $1$-meter diameter half sphere, (2) xyz: the camera moves along the \mbox{x-y-z} axes, (3) rpy: the camera rotates over roll, pitch and yaw axes, and (4) static: the camera is kept static manually.

We use the absolute trajectory RMSE as the error metric for our experiments, as proposed by Sturm \textit{et~al.} \cite{sturm2012benchmark}. 

\begin{figure} [b!]
\centering
\includegraphics[width=0.97\linewidth]{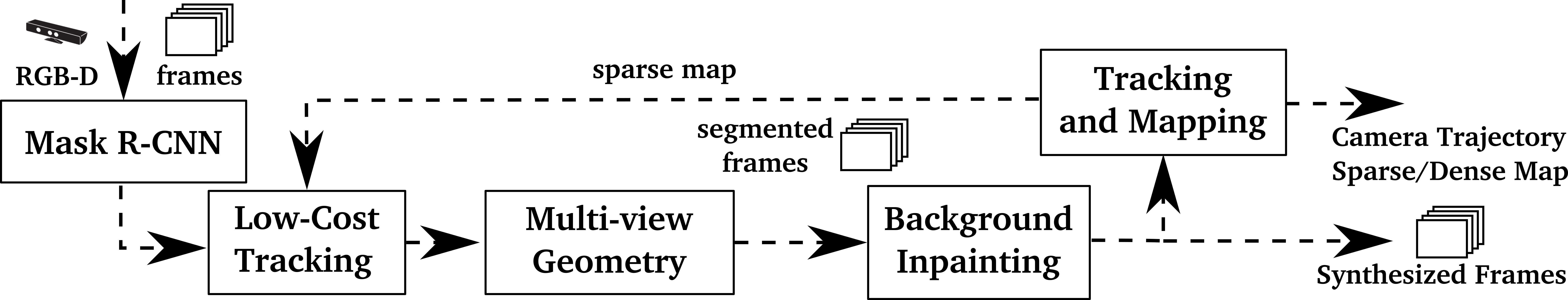}
\vspace*{-0.3\baselineskip}
\caption{\label{fig:RGBDdiagram} Block diagram of RGB-D DynaSLAM (N+G+BI).}
\end{figure}

The results of different variations of our system for six sequences within this dataset are shown in \mbox{Table \ref{tab:intern}}. Firstly, DynaSLAM (N) is the system in which only \mbox{Mask R-CNN} segments out the \textit{a priori} dynamic objects.
Secondly, in DynaSLAM (G) the dynamic objects have been only detected with the multi-view geometry method based on depth changes. Thirdly, DynaSLAM (N+G) stands for the system in which the dynamic objects have been detected combining both the geometrical and deep learning approaches. Finally, we have considered interesting to analyze the system shown in \mbox{Fig. \ref{fig:RGBDdiagram}}. 
In this case (N+G+BI), the background inpainting stage (BI) is done before the tracking and mapping. The motivation for this experiment is that, if the dynamic areas are inpainted with the static content, the system can work as a SLAM system under the staticity assumption using the inpainted images. 
In this proposal, the ORB features extractor algorithm works both in the real and reconstructed areas of the frames, finding matches with the keypoints of the previously processed keyframes.


\setlength{\tabcolsep}{4pt}
\begin{table} [t!]
\begin{tabularx}{\linewidth}{@{}l*{5}{C}}
\toprule
Sequence & DynaSLAM \mbox{(N)} & DynaSLAM \mbox{(G)} & DynaSLAM \mbox{(N+G)} & DynaSLAM \mbox{(N+G+BI)} \\
\midrule
$w\_halfsphere$ & \textbf{0.025} & 0.035 & \textbf{0.025} & 0.029\\
$w\_xyz$ & \textbf{0.015} & 0.312 & \textbf{0.015} & \textbf{0.015} \\
$w\_rpy$ & 0.040 & 0.251 & \textbf{0.035} & 0.136 \\
$w\_static$ & 0.009 & 0.009 & \textbf{0.006} & 0.007\\
\addlinespace
$s\_halfsphere$ & \textbf{0.017} & 0.018 & \textbf{0.017} & 0.025 \\
$s\_xyz$ & 0.014 & \textbf{0.009} & 0.015 & 0.013 \\
\bottomrule
\end{tabularx}
\vspace*{-0.5\baselineskip}
\caption{\label{tab:intern} Absolute trajectory RMSE [m] for several variants of DynaSLAM (RGB-D).}
\end{table}

According to Table \ref{tab:intern}, the system (N+G) that uses learning and geometry is the most accurate one in most sequences. The improvement over (N) comes from the segmentation of \emph{movable} objects and refinement of the dynamic segments. The system (G) has higher error because it needs motion and its segmentation is only accurate after a small delay, during which the dynamic content introduces some error in the estimation.

Adding the background inpainting stage (BI) before the localization of the camera (Fig. \ref{fig:RGBDdiagram}) usually leads to less accuracy in the tracking. 
The reason is that the background reconstruction is strongly correlated with the camera poses. 
Hence, for sequences with purely rotational motion (\textit{rpy}, \textit{halfsphere}), the estimated camera poses have a greater error and lead to a non-accurate background reconstruction. The background inpainting stage (BI) should be done therefore once the tracking stage is finished (Fig. \ref{fig:diagram}). The main accomplishment of the background reconstruction is seen in the synthesis of the static images (Fig. \ref{fig:visual results}) for applications such as virtual reality or cinematography.  The DynaSLAM results shown from now on are from the best variant, that is, (N+G). 

\begin{figure} [b!]
\centering 
\input{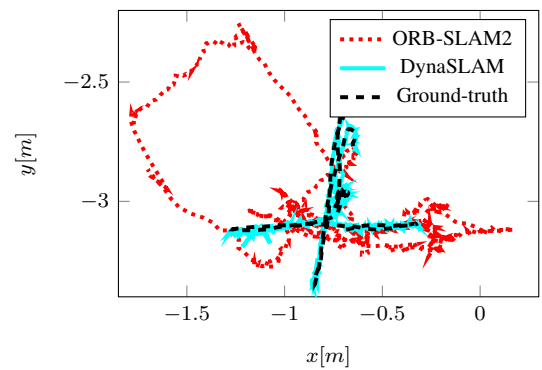}
\caption{\label{fig:traj} Ground truth and trajectories estimated by DynaSLAM and ORB-SLAM2 in the TUM sequence $fr3/walking\_xyz$.}
\end{figure}

Table \ref{tab:RGB-D} shows our results on the same  sequences, compared against \mbox{RGB-D} ORB-SLAM2. Our method outperforms ORB-SLAM2 in highly dynamic scenarios ($walking$), reaching an error similar to that of the original \mbox{RGB-D} ORB-SLAM2 system in static scenarios. In the case of low-dynamic scenes ($sitting$) the tracking results are slightly worse because the tracked keypoints find themselves further than those belonging to dynamic objects. Albeit, DynaSLAM's map does not contain the dynamic objects that appear along the sequence. Fig. \ref{fig:traj} shows an example of the estimated trajectories of DynaSLAM and ORB-SLAM2, compared to the \mbox{ground-truth}.

\setlength{\tabcolsep}{4pt}
\begin{table} [h!]
\begin{tabularx}{\linewidth}{@{}l*{5}{C}}
\toprule
Sequence & \mbox{ORB-SLAM2} \mbox{(RGB-D) \cite{mur2017orb}}  & \multicolumn{3}{c}{DynaSLAM (N+G) \mbox{(RGB-D)}}\\
 & median & median & min & max \\
\midrule
$w\_halfsphere$ & 0.351 & \textbf{0.025} & 0.024 & 0.031 \\
$w\_xyz$ & 0.459 & \textbf{0.015} & 0.014 & 0.016 \\
$w\_rpy$ & 0.662 & \textbf{0.035} & 0.032 & 0.038 \\
$w\_static$ & 0.090 & \textbf{0.006} & 0.006 & 0.008 \\
\addlinespace
$s\_halfsphere$ & 0.020 & \textbf{0.017} & 0.016 & 0.020 \\
$s\_xyz$ & \textbf{0.009} & 0.015 & 0.013 & 0.015 \\
\bottomrule
\end{tabularx}
\vspace*{-0.5\baselineskip}
\caption{\label{tab:RGB-D} Comparison of the RMSE of ATE [m] of DynaSLAM against ORB-SLAM2 for \mbox{RGB-D} cameras. To account for the non-deterministic nature of the system, we show the median, minimum and maximum error of ten runs.}
\end{table}

{\begin{table*} [h!]
\begin{tabularx}{\linewidth}{@{}l*{11}{C}}
\toprule
Sequence & \mbox{Depth Edge} & \multicolumn{3}{c}{Motion Segmentation DSLAM \cite{wang2014motion}} & \multicolumn{3}{c}{Motion Removal DVO-SLAM \cite{sun2017improving}} & \multicolumn{3}{c}{DynaSLAM (N+G) (\mbox{RGB-D})} \\
&  \mbox{SLAM \cite{li2017rgb}}& \multicolumn{3}{c}{} & \multicolumn{3}{c}{} & \multicolumn{3}{c}{} \\
 &  & \mbox{w/o Motion} Detection & \mbox{w/ Motion} Detection & Improvement w/ MD & \mbox{w/o Motion} Detection & \mbox{w/ Motion} Detection & Improvement w/ MD & \mbox{w/o Motion} Detection & \mbox{w/ Motion} Detection & Improvement w/ MD \\
  & [m] & [m] & [m] & [\%] & [m] & [m] & [\%] & \cite{mur2017orb} [m] & [m] & [\%]\\
\midrule
$w\_half$ & \multicolumn{1}{|c|}{0.049} & 0.116 & 0.055 & \multicolumn{1}{c|}{52.59\%} & 0.529 & 0.125 & \multicolumn{1}{c|}{76.32\%} & 0.351 & \textbf{0.025} & \textbf{92.88\%} \\
$w\_xyz$ & \multicolumn{1}{|c|}{0.060} & 0.202 & 0.040 & \multicolumn{1}{c|}{80.20\%} & 0.597 & 0.093 & \multicolumn{1}{c|}{84.38\%} & 0.459 & \textbf{0.015} & \textbf{96.73\%} \\ 
$w\_rpy$ & \multicolumn{1}{|c|}{0.179} & 0.515 & 0.076 & \multicolumn{1}{c|}{85.24\%} & 0.730 & 0.133 & \multicolumn{1}{c|}{81.75\%} & 0.662 & \textbf{0.035} & \textbf{94.71\%} \\
$w\_stat$ & \multicolumn{1}{|c|}{0.026} & 0.470 & 0.024 & \multicolumn{1}{c|}{\textbf{94.89\%}} & 0.212 & 0.066 & \multicolumn{1}{c|}{69.06\%} & 0.090 & \textbf{0.006} & 93.33\% \\
\addlinespace
$s\_half$ & \multicolumn{1}{|c|}{0.043} & - & - & \multicolumn{1}{c|}{-} & 0.062 & 0.047 & \multicolumn{1}{c|}{\textbf{23.70\%}} & 0.020 & \textbf{0.017} & 15.00\% \\
$s\_xyz$ & \multicolumn{1}{|c|}{0.040} & - & - & \multicolumn{1}{c|}{-} & 0.051 & 0.048 & \multicolumn{1}{c|}{\textbf{4.55\%}} & 0.009 & \textbf{0.015} & X \\ 
\bottomrule
\end{tabularx}
\vspace*{-0.5\baselineskip}
\caption{\label{tab:StateArt} Absolute trajectory RMSE [m] of DynaSLAM against state-of-the-art \mbox{RGB-D} SLAM systems in dynamic scenes. To evaluate the effectiveness of the specific module addressing  dynamic content, we report the improvement with respect to the original SLAM systems (w/o Motion Detection). Our results are estimated using \mbox{Mask R-CNN} and multi-view geometry.}
\end{table*}

Table \ref{tab:StateArt} shows a comparison between our system and several state-of-the-art \mbox{RGB-D} SLAM systems designed for dynamic environments. 
In account for the effectiveness of our and the state-of-the-art approaches for motion detection (independently of the utilized SLAM system), we also show the respective improvement values against the original SLAM system used in every case. 
DynaSLAM significantly outperforms all of them in all sequences (both high and low dynamic ones). The error is, in general, around 1-2 cm, similar to that of the state of the art in static scenes. 
Our motion detection approach also outperforms the other methods.

ORB-SLAM, the monocular version of ORB-SLAM2, is generally more accurate than the \mbox{RGB-D} one in dynamic scenes, due to their different initialization algorithms. \mbox{RGB-D} ORB-SLAM2 is initialized and starts the tracking from the very first frame, and hence dynamic objects can introduce errors. ORB-SLAM delays the initialization until there is parallax and consensus using the staticity assumption. Hence, it does not track the camera for the full sequence, sometimes missing a substantial part of it, or even not initializing.


Table \ref{tab:monoTUM} shows the tracking results and percentage of the tracked trajectory for ORB-SLAM and DynaSLAM (monocular) in the TUM dataset. The initialization in DynaSLAM is always quicker than that of ORB-SLAM. In fact, in highly dynamic sequences, ORB-SLAM initialization only occurs when the moving objects disappear from the scene. In conclusion, although the accuracy of DynaSLAM is slightly lower, it succeeds in bootstrapping the system with dynamic content and producing a map without such content (see \mbox{Fig. \ref{fig:overview}}), to be re-used for long-term applications. 
The reason why DynaSLAM is slightly less accurate is that the estimated trajectory is longer, and there is therefore room for accumulating errors.

\setlength{\tabcolsep}{4pt}
\begin{table} [h!]
\begin{tabularx}{\linewidth}{@{}l*{5}{C}}
\toprule
Sequence & \multicolumn{2}{c}{ORB-SLAM} & \multicolumn{2}{c}{DynaSLAM} \\
 & \multicolumn{2}{c}{\cite{mur2017orb}} & \multicolumn{2}{c}{(Monocular)} \\
\addlinespace
 & \mbox{ATE [m]} & \mbox{\% Traj} & \mbox{ATE [m]} & {\mbox{\% Traj}} \\
\midrule
$fr3/walking\_halfsphere$ & \textbf{0.017} & 87.16 & 0.021 & \textbf{97.84} \\
$fr3/walking\_xyz$ & \textbf{0.012} & 57.63 & 0.014 & \textbf{87.37} \\
\addlinespace
$fr2/desk\_with\_person$ & \textbf{0.006} & 95.30 & 0.008 & \textbf{97.07} \\
$fr3/sitting\_xyz$ & \textbf{0.007} & 91.44 & 0.013 & \textbf{100.00}\\
\bottomrule
\end{tabularx}
\vspace*{-0.5\baselineskip}
\caption{\label{tab:monoTUM} Absolute trajectory RMSE [m] and percentage of successfully tracked trajectory for both ORB-SLAM and DynaSLAM (monocular).}
\end{table}

\subsection{KITTI Dataset}
The KITTI Dataset \cite{geiger2013vision} contains stereo sequences recorded from a car in urban and highway environments. Table \ref{tab:stereo} shows our results in the eleven training sequences, compared against stereo ORB-SLAM2. We use two different metrics, the absolute trajectory RMSE proposed in \cite{sturm2012benchmark}, and the average relative translation and rotation errors, proposed in \cite{geiger2013vision}. Table \ref{tab:monoKITTI} shows the results in the same sequences for the monocular variants of ORB-SLAM and DynaSLAM.

\setlength{\tabcolsep}{4pt}
\begin{table} [h!]
\begin{tabularx}{\linewidth}{@{}l*{7}{C}}
\toprule
Sequence & \multicolumn{3}{c}{ORB-SLAM2 (Stereo) \cite{mur2017orb}} & \multicolumn{3}{c}{DynaSLAM (Stereo)} \\
\addlinespace
& RPE & RRE & ATE & RPE & RRE & ATE  \\ 
& [\%] & [$^\circ$/100m] & [m] & [\%] & [$^\circ$/100m] & [m]\\
\midrule
KITTI 00 & \textbf{0.70} & \textbf{0.25} & \multicolumn{1}{c|}{\textbf{1.3}} & 0.74 & 0.26 & 1.4 \\
KITTI 01 & \textbf{1.39} & \textbf{0.21} & \multicolumn{1}{c|}{10.4} & 1.57 & 0.22 & \textbf{9.4} \\
KITTI 02 & \textbf{0.76} & \textbf{0.23} & \multicolumn{1}{c|}{\textbf{5.7}} & 0.80 & 0.24 & 6.7 \\
KITTI 03 & 0.71 & 0.18 & \multicolumn{1}{c|}{0.6} & \textbf{0.69} & 0.18 & 0.6 \\
KITTI 04 & 0.48 & 0.13 & \multicolumn{1}{c|}{0.2} & \textbf{0.45} & \textbf{0.09} & 0.2 \\
KITTI 05 & 0.40 & 0.16 & \multicolumn{1}{c|}{0.8} & 0.40 & 0.16 & 0.8 \\
KITTI 06 & 0.51 & \textbf{0.15} & \multicolumn{1}{c|}{0.8} & \textbf{0.50} & 0.17 & 0.8  \\
KITTI 07 & \textbf{0.50} & \textbf{0.28} & \multicolumn{1}{c|}{0.5} & 0.52 & 0.29 & 0.5 \\
KITTI 08 & 1.05 & 0.32 & \multicolumn{1}{c|}{3.6} & 1.05 & 0.32 & \textbf{3.5} \\
KITTI 09 & \textbf{0.87} & \textbf{0.27} & \multicolumn{1}{c|}{3.2} & 0.93 & 0.29 & \textbf{1.6} \\
KITTI 10 & \textbf{0.60} & \textbf{0.27} & \multicolumn{1}{c|}{\textbf{1.0}} & 0.67 & 0.32 & 1.2 \\
\bottomrule
\end{tabularx}
\vspace*{-0.5\baselineskip}
\caption{\label{tab:stereo} Comparison of the RMSE of the ATE [m], the average of the RPE [$\%$] and the RRE [$^\circ/100m$] of DynaSLAM against ORB-SLAM2 system for stereo cameras.}
\end{table}

\setlength{\tabcolsep}{4pt}
\begin{table} [h!]
\begin{tabularx}{\linewidth}{@{}l*{3}{C}}
\toprule
Sequence & ORB-SLAM \cite{mur2017orb} & DynaSLAM (Monocular) \\
\midrule
KITTI 00 & \textbf{5.33} & 7.55 \\
KITTI 02 & \textbf{21.28} & 26.29 \\
KITTI 03 & \textbf{1.51} & 1.81 \\
KITTI 04 & 1.62 & \textbf{0.97} \\
KITTI 05 & 4.85 & \textbf{4.60} \\
KITTI 06 & \textbf{12.34} & 14.74 \\
KITTI 07 & \textbf{2.26} & 2.36\\
KITTI 08 & 46.68 & \textbf{40.28}\\
KITTI 09 & 6.62 & \textbf{3.32} \\
KITTI 10 & 8.80 & \textbf{6.78}\\
\bottomrule
\end{tabularx}
\vspace*{-0.5\baselineskip}
\caption{\label{tab:monoKITTI} Absolute trajectory RMSE [m] for ORB-SLAM and DynaSLAM (monocular).}
\end{table}

Note that the results are similar in both the monocular and stereo cases, but the former is more sensitive to dynamic objects and therefore to the additions in DynaSLAM. In some sequences the accuracy of the tracking is improved when not using features belonging to \textit{a priori} dynamic objects, \textit{i.e.}, cars, bicycles, \textit{etc}. An example of this would be the sequences KITTI 01 and KITTI 04, in which all vehicles that appear are moving. In the sequences in which most of the recorded cars and vehicles are parked (hence static), the absolute trajectory RMSE is usually bigger since the keypoints used for tracking are more distant and usually belong to low-texture areas (KITTI 00, KITTI 02, KITTI 06). However, the loop closure and relocalization algorithms work more robustly since the resulting map only contains structural objects, \textit{i.e.}, the map can be re-used and work in long-term applications. 

As future work, it is interesting to make a distinction between those \textit{movable} and \textit{moving} objects, by using only RGB information. If a car is detected by the CNN (\textit{movable}) but is not currently moving, its corresponding keypoints should be used for the local tracking, but should not be in the map. 

\subsection{Timing Analysis}
To complete the evaluation of our proposal, Table~\ref{tab:Time} shows the average computational time for its different stages. Note that DynaSLAM is not optimized for real-time operation. However, its capability for creating life-long maps of the static scene content are also relevant for running on offline mode. 

\setlength{\tabcolsep}{4pt}
\begin{table} [h!]
\begin{tabularx}{\linewidth}{@{}l*{4}{C}}
\toprule
Sequence & Low-Cost Tracking [ms] & Multi-view Geometry [ms] & Background Inpainting [ms]\\ 
\midrule
$w\_halfsphere$ & 1.69 & 333.68 & 208.09 \\
$w\_rpy$ & 1.59 & 235.98 & 183.56 \\
\bottomrule
\end{tabularx}
\vspace*{-0.5\baselineskip}
\caption{\label{tab:Time} DynaSLAM average computational time [ms].}
\end{table}

Mur \textit{et~al.} show real-time results for and ORB-SLAM2 \cite{mur2017orb}. 
He \textit{et al.} \cite{he2017mask} report that \mbox{Mask R-CNN}} runs at $195$ ms per image on a Nvidia Tesla M40 GPU.

The addition of the multi-view geometry stage is an additional slowdown, due mainly to the region growth algorithm. The background inpainting also introduces a delay, which is another reason why it should be done after the tracking and mapping stage, as it has been shown in Fig. \ref{fig:diagram}.


\section{CONCLUSIONS}
\label{sec:conclusions}

We have presented a visual SLAM system that, building on ORB-SLAM, adds a motion segmentation approach that makes it robust in dynamic environments for monocular, stereo and \mbox{RGB-D} cameras.
Our system accurately tracks the camera and creates a static and therefore reusable map of the scene. In the \mbox{RGB-D} case, DynaSLAM is capable of obtaining the synthetic RGB frames with no dynamic content and with the occluded background inpainted, as well as their corresponding synthesized depth frames, which might be together very useful for virtual reality applications. We include a video showing the potential of DynaSLAM
\footnote{\href{https://youtu.be/EabI_goFmQs}{{\tt https://youtu.be/EabI\_goFmQs}}}. 

The comparison against the state of the art shows that DynaSLAM achieves in most cases the highest accuracy. 

In the TUM Dynamic Objects dataset, DynaSLAM is currently the best \mbox{RGB-D} SLAM solution. 
In the monocular case, our accuracy is similar to that of ORB-SLAM, obtaining however a static map of the scene with an earlier initialization.

In the KITTI dataset DynaSLAM is slightly less accurate than monocular and stereo ORB-SLAM, except for those cases in which dynamic objects represent an important part of the scene. However, our estimated map only contains structural objects and can therefore be re-used in long-term applications.

Future extensions of this work might include, among others, real-time performance, an RGB-based motion detector, or a more realistic appearance of the synthesized RGB frames by using a more elaborate inpainting technique, \textit{e.g.}, the one used by Pathak \textit{et al.} \cite{pathak2016context} by the use of GANs.


\bibliographystyle{ieeetr}
\bibliography{ref}

\end{document}